\documentclass[runningheads]{llncs}
\usepackage{graphicx}
\usepackage{amsmath,amssymb} 
\usepackage{color}

\usepackage{times}
\usepackage{epsfig}
\usepackage{mathtools}
\usepackage{color, colortbl}

\usepackage[ruled,vlined,linesnumbered]{algorithm2e}

\usepackage[utf8]{inputenc}
\usepackage{siunitx}
\graphicspath{ {img/} }
\usepackage[caption=false]{subfig}
\usepackage{enumitem}
\usepackage{booktabs}

\usepackage{array}
\newcolumntype{P}[1]{>{\centering\arraybackslash}p{#1}}
\newcolumntype{M}[1]{>{\centering\arraybackslash}m{#1}}

\DeclarePairedDelimiter\norm{\lVert}{\rVert}%

\newcommand{\mybold}[1]{\boldsymbol{\mathbf{#1}}}

\newcommand{\bb}{\mybold{b}}
\newcommand{\cb}{\mybold{c}}
\newcommand{\fb}{\mybold{f}}

\newcommand{\hb}{\mybold{h}}
\newcommand{\ib}{\mybold{i}}
\newcommand{\ob}{\mybold{o}}

\newcommand{\mb}{\mybold{m}}
\newcommand{\vb}{\mybold{v}}

\newcommand{\Wb}{\mybold{W}}

\newcommand{\figref}[1]{Fig.~\ref{#1}}
\newcommand{\secref}[1]{Sec.~\ref{#1}}
\newcommand{\tabref}[1]{Table~\ref{#1}}

\newcommand{\mmd}{\text{MMD}}
\newcommand{\lnlstm}{\text{LNLSTM}}
\newcommand{\fone}{$F_1$}

\newcommand{\RemoveAboveCaption}{-7.5pt}
\newcommand{\RemoveBelowCaption}{0pt}

\usepackage{array}
\newcolumntype{P}[1]{>{\centering\arraybackslash}p{#1}}
\newcolumntype{M}[1]{>{\centering\arraybackslash}m{#1}}

\usepackage{xspace}
\makeatletter
\def\eg{\emph{e.g}.\@\xspace} 
\def\ie{\emph{i.e}.\@\xspace}

\def\etal{\emph{et al}.\@\xspace}
\makeatother

\begin{document}
\title{Human Motion Analysis with Deep Metric Learning} 

\titlerunning{Human Motion Analysis with Deep Metric Learning}
%
\author{Huseyin Coskun\inst{1,*} \and David Joseph Tan\inst{1,2,*} \and Sailesh Conjeti\inst{1}\and Nassir Navab\inst{1,2} \and Federico Tombari\inst{1,2}}
%
\authorrunning{H. Coskun, D. J. Tan, S. Conjeti, N. Navab \and F. Tombari}
%

\institute{Technische Universit\"at M\"unchen, Germany \and
Pointu3D GmbH, Germany}
\maketitle              
\begin{abstract}


Effectively measuring the similarity between two human motions is necessary for several computer vision tasks such as gait analysis, person identification and action retrieval. Nevertheless, we believe that traditional approaches such as L2 distance or Dynamic Time Warping based on hand-crafted local pose metrics fail to appropriately capture the semantic relationship across motions and, as such, are not suitable for being employed as metrics within these tasks. 
This work addresses this limitation by means of a triplet-based deep metric learning specifically tailored to deal with human motion data, in particular with the problem of varying input size and computationally expensive hard negative mining due to motion pair alignment. 
Specifically, we propose (1)~a novel metric learning objective based on a triplet architecture and Maximum Mean Discrepancy; as well as, (2)~a novel deep architecture based on attentive recurrent neural networks. 
One benefit of our objective function is that it enforces a better separation within the learned embedding space of the different motion categories by means of the associated distribution moments. At the same time, our attentive recurrent neural network allows processing varying input sizes to a fixed size of embedding while learning to focus on those motion parts that are semantically distinctive. 
Our experiments on two different datasets demonstrate significant improvements over conventional human motion metrics.


\end{abstract}

{\let\thefootnote\relax\footnotetext{* Equal contribution}}

\begin{figure} [t]
	\centering
	\includegraphics[width=1.00\textwidth]{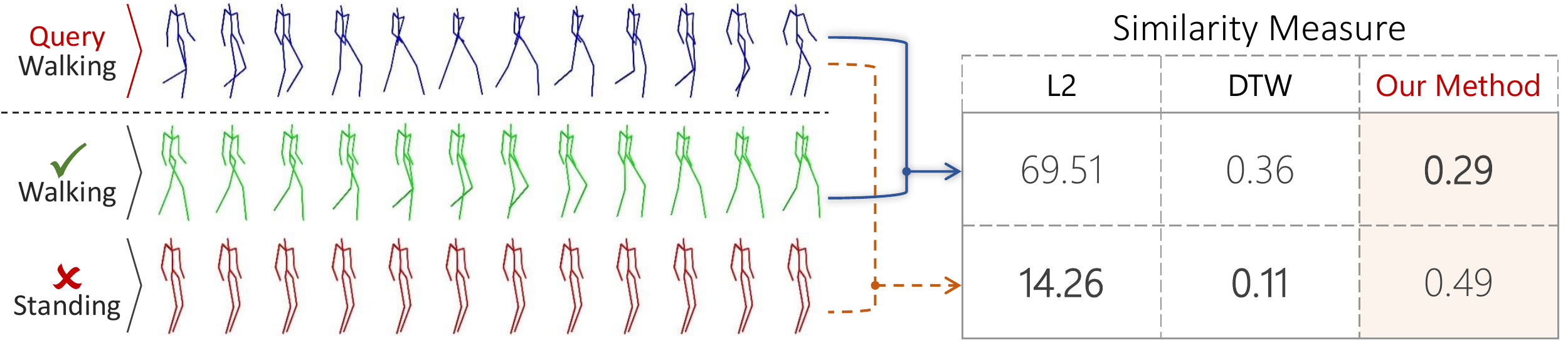}
	\setlength{\abovecaptionskip}{\RemoveAboveCaption}
	\setlength{\belowcaptionskip}{\RemoveBelowCaption}
	\caption{When asked to measure the similarity to a query sequence (``Walking", top), both the L2 and the DTW measures judge the unrelated sequence (``Standing", bottom) as notably more similar compared to a semantically correlated one (``Walking", middle). Conversely, our learned metric is able to capture the contextual information and measure the similarity correctly with respect to the given labels.}
	\label{fig:teaser}
\end{figure}

\section{Introduction}

In image-based human pose estimation, the similarity between two predicted poses can be precisely assessed through conventional approaches that either evaluate the distance between corresponding joint locations~\cite{Chu_2017_CVPR,newell2016stacked,Yang_2017_ICCV} or the average difference of corresponding joint angles~\cite{mehta2017vnect,sun2017compositional}. 
Nevertheless, when human poses have to be compared across a temporal set of frames, the assessment of the similarity between two sequences of poses or motion becomes a non-trivial problem. 
Indeed, human motion typically evolves in a different manner on different sequences, which means that specific pose patterns tend to appear at different time instants on sequences representing the same human motion: see, \eg, the first two sequences in \figref{fig:teaser}, which depict two actions belonging to the same class. 
Moreover, these sequences result also in varying length (\ie, a different number of frames), this making the definition of a general similarity measure more complicated. 
Nevertheless, albeit challenging, estimating the similarity between human poses across a sequence is a required step in human motion analysis tasks such as action retrieval and recognition, gait analysis and motion-based person identification. 


Conventional approaches deployed to compare human motion sequences are based on estimating the L2 displacement error~\cite{Martinez_2017_CVPR} or Dynamic Time Warping (DTW)~\cite{vintsyuk1968speech}. 
Specifically, the former computes the squared distance between corresponding joints in the two sequences at a specific time $t$. As shown by Martinez~\etal~\cite{Martinez_2017_CVPR}, such measure tends to disregard the specific motion characteristics, since a constant pose repeated over a sequence might turn out to be a better match to a reference sequence than a visually similar motion with a different temporal evolution. 
On the other hand, DTW tries to alleviate this problem by warping the two sequences via compressions or expansions so to maximize the matching between local poses. 
Nevertheless, DTW can easily fail in appropriately estimating the similarity when the motion dynamic in terms of peaks and plateaus exhibits small temporal variations, as shown in \cite{keogh2001derivative}. 
As an example, \figref{fig:teaser} illustrates a typical failure case of DTW when measuring the similarity among three human motions. 
Although the first two motions are visually similar to each other while the third one is unrelated to them, DTW estimates a smaller distance between the first and the third sequence. 
In general, neither the DTW nor the L2 metrics can comprehensively capture the semantic relationship between two sequences since they disregard the contextual information (in the temporal sense), this limiting their application in the aforementioned scenarios. 

The goal of this work is to introduce a novel metric for estimating the similarity between two human motion sequences. 
Our approach relies on deep metric learning that uses a neural network to map high-dimensional data to a low-dimensional embedding~\cite{rippel2016metric,schroff2015facenet,sohn2016improved,zhang2017learning}. 
In particular, our first contribution is to design an approach so to map semantically similar motions over nearby locations in the learned embedding space. This allows the network to express a similarity measure that strongly relies on the motion's semantic and contextual information. 
To this end, we employ a novel objective function based on the Maximum Mean Discrepancy (MMD)~\cite{gretton2012kernel}, which enforces motions to be embedded based on their distribution moments. The main advantage with respect to standard triplet loss learning is represented by the fact that our approach, being based on distributions and not samples, does not require hard negative mining to converge, which is computationally expensive since finding hard negatives in a human motion datasets requires the alignment of sequence pairs, which has an $O(n^2)$ complexity ($n$ being the sequence length).
As our second main contribution, we design a novel deep learning architecture based on attentive recurrent neural networks (RNNs) which exploits attention mechanisms to map an arbitrary input size to a fixed sized embedding while selectively focusing on the semantically descriptive parts of the motion. 

One advantage of our approach is that, unlike DTW, we do not need any explicit synchronization or alignment of the motion patterns appearing on the two sequences, since motion patterns are implicitly and semantically matched via deep metric learning. 
In addition, our approach can naturally deal with varied size input thanks to the use of the recurrent model, while retaining the distinctive motion patterns by means of the attention mechanism. An example is shown in  \figref{fig:teaser}, comparing our similarity measure to DTW and L2.
We validate the usefulness of our approach for the tasks of action retrieval and motion-based person identification on two publicly available benchmark datasets. The proposed experiments demonstrate significant improvements over conventional human motion similarity metrics.

\section{Related work} \label{sec:rel_work}

In recent literature, image-based deep metric learning has been extensively studied. However, just a few works focused on metric learning for time-series data, in particular human motion. 
Here, we first review metric learning approaches for human motion, then follow up with recent improvements in deep metric learning.

\paragraph{Metric learning for time series and human motion.} 

We first review metric learning approaches for time series, then focus only on works related on human motion analysis. 
Early works on metric learning for time series approaches measure the similarity in a two steps process \cite{berndt1994using,cuturi2007kernel,ratanamahatana2004making}. 
First, the model determines the best alignment between two time series, then it computes the distance based on the aligned series. 
Usually, the model finds the best alignment by means of the DTW measure, first by considering all possible alignments, then ranking them based on hand-crafted local metric. 
These approaches have two main drawbacks: first, the model yields an $O(n^2)$ complexity; secondly, and most importantly, the local metric can hardly capture relationship in high dimensional data. 
In order to overcome these drawbacks, Mei~\etal~\cite{mei2016learning} propose to use LogDet divergence to learn a local metric that can capture the relationship in high dimensional data. 
Che~\etal~\cite{che2017decade} overcome the hand crafted local metric problem by using a feed-forward network to learn local similarities. 
Although the proposed approaches~\cite{che2017decade,mei2016learning} learn to measure the similarity between two given time series at time $t$, the relationship between two time steps is discarded. Moreover, finding the best alignment requires to search for all possible alignments. 
To address these problems, recent work focused on determining a low dimensional embedding to measure the distance between time series. To this goal, Pei~\etal~\cite{pei2016modeling} and Zheng~\etal~\cite{zheng2015convolutional} used a Siamese network which learns from pairs of inputs. While Pei~\etal~\cite{pei2016modeling} trained their network by minimizing the binary cross entropy in order to predict whether the two given time series belong to the same cluster or not, Zheng~\etal~\cite{zheng2015convolutional} propose to minimize a loss function based on the Neighbourhood Component Analysis (NCA) \cite{roweis2004neighbourhood}. The main drawback of these approaches is that the siamese architecture learns the embedding by considering only the relative distances between the provided input pairs.

As for metric learning for human motion analysis, they mostly focus on directly measuring the similarity between corresponding poses along the two sequences. 
Lopez~\etal~\cite{lopez2012metric} proposed a model based on \cite{davis2007information} to learn a distance metric for two given human poses, while aligning the motions via Hidden Markov Models (HMM) \cite{eddy1996hidden}. Chen~\etal~\cite{chen2011learning} proposed a semi-supervised learning approach built on a hand-crafted geometric pose feature and aligned via DTW. 
By considering both the pose similarity and the pose alignment in learning, Yin~\etal~\cite{yin2016deep} proposed to learn pose embeddings with an auto-encoder trained with an alignment constraint. Notably, this approach requires an initial alignment based on DTW. The main drawback of these approaches is that their accuracy relies heavily on the accurate motion alignment provided by HMM or DTW, which is computationally expensive to obtain and prone to fail in many cases. Moreover, since the learning process considers only single poses, they lack at capturing the semantics of the entire motion.

\paragraph{Recent improvements in deep metric learning.}

Metric learning with deep networks started with Siamese architectures that minimize the contrastive loss~\cite{chopra2005learning,hadsell2006dimensionality}. 
Schroff~\etal~\cite{schroff2015facenet} suggest using a triplet loss to learn the embeddings on facial recognition and verification, showing that it performs better than contrastive loss to learn features. 
%
%
Since they conduct hard-negative mining, 
when the training set and the number of different categories increase, searching for hard-negatives become computationally inefficient. 
Since then, research mostly focus on carefully constructing batches and using all samples in the batch. 
Song~\etal~\cite{song2016deep} proposed the lifted loss for training, so to use all samples in a batch. 
In \cite{sohn2016improved}, they further developed the idea and propose an $n$-pair loss that uses all negative samples in a batch. Other triplet-based approaches are \cite{HardNet2017,tian2017l2}.
In \cite{rippel2016metric}, the authors show that minimizing the loss function computed on individual pairs or triplets does not necessarily enforce the network to learn features that represent contextual relations between clusters. 
Magnet Loss~\cite{rippel2016metric} address some of these issues by learning features that compare the distributions rather than the samples. 
Each cluster distribution is represented by the cluster centroid obtained via $k$-means algorithm. 
A shortcoming of this approach is that computing cluster centers requires to interrupt training, this slowing down the process. 
Proxy-NCA~\cite{movshovitz2017no} tackle this issue by designing a network architecture that learns the cluster centroids in an end-to-end fashion, this avoiding interruptions during training. 
Both the Magnet Loss and the Proxy-NCA use the NCA~\cite{roweis2004neighbourhood} loss to compare the samples. Importantly, they both represent distributions with cluster centroids which do not convey sufficient contextual information of the actual categories, and require to set a pre-defined number of clusters. 
In contrast, we propose to use a loss function based on MMD~\cite{gretton2012kernel}, which relies on distribution moments that do not need to explicitly determine or learn cluster centroids.

\section{Metric learning on human motion} 
\label{sec:method-MMD-NCA}



The objective is to learn an embedding for human motion sequences, such that the similarity metric between two human motion sequences $X := \{x_{1}, x_{2}, ..., x_{n}\}$ and $Y := \{y_{1}, y_{2}, ..., y_{m}\}$ (where $x_{t}$ and $y_{t}$ represent the poses at time $t$) can be expressed directly as the squared Euclidean distance in the embedding space. 
Mathematically, this can be written as
\begin{equation}
d(f(X), f(Y)) = \left\Vert f(X)- f(Y) \right\Vert^2
\end{equation} 
where $f(\cdot)$ is the learned embedding function that maps a varied-length motion sequence to a point in a Euclidean space, and $d(\cdot,\cdot)$ is the squared Euclidean distance. 
The challenge of metric learning is to find a motion embedding function $f$ such that the distance $d(f(X), f(Y))$ should be inversely proportional to the similarity of the two sequences $X$ and $Y$. 
In this paper, we learn $f$ by means of a deep learning model trained with a loss function (defined in \secref{sec:loss_function}) which is derived from the integration of MMD with a triplet learning paradigm. In addition, its architecture (described in \secref{sec:method-network}) is based on an attentive recurrent neural network.

\section{Loss function}
\label{sec:loss_function}


Following the standard deep metric learning approach, we model the embedding function $f$ by minimizing the distance $d(f(X), f(Y))$ when $X$ and $Y$ belong to the same category, while maximizing it otherwise. A conventional way of learning $f$ would be to train a network with the contrastive loss~\cite{chopra2005learning,hadsell2006dimensionality}
\begin{equation}
\mathcal{L}_\text{contrastive}= (r) \frac{1}{2} d
+  (1-r) \frac{1}{2}  [ \max(0, \alpha_\text{margin}-d) ]^2
\label{eq:contrastive}
\end{equation} 
where $r \in \{1,0\}$ indicates whether $X$ and $Y$ are from the same category or not, and $\alpha_\text{margin}$ defines the margin between different category samples. 
During training, the contrastive loss penalizes those cases where different category samples are closer than $\alpha_\text{margin}$ and when the same category samples have a distance greater than zero. 
This equation shows that the contrastive loss only takes into account pairwise relationships between samples, thus only partially exploiting relative relationships among categories. Conversely, triplet learning better exploit such relationships by taking into account three samples at the same time, where the first two are from the same category while the third is from a different one. 
Notably, it has been shown that exploiting relative relationships among categories play a fundamental role in terms of the quality of the learned embedding~\cite{schroff2015facenet,zhang2017learning}. 
The triplet loss enforces embedding samples from the same category with a given margin distance with respect to samples from a different category.
If we denote the three human motion samples as $X$, $X^+$ and $X^-$, the commonly used ranking loss~\cite{schultz2004learning} takes the form of 
\begin{equation}
\mathcal{L}_\text{triplet}=\max(0, ~\norm{f(X)-f(X^+)}^2- \norm{f(X)-f(X^-)}^2+\alpha_\text{margin})
\label{eq:triplet}
\end{equation} 
where $X$ and $X^+$ represent the motion samples from the same category and $X^-$ represents the sample from a different category.
In literature $X$, $X^+$, and $X^-$ are often referred to as anchor, positive, and negative samples, respectively~\cite{rippel2016metric,schroff2015facenet,sohn2016improved,zhang2017learning}. 

However, one of the main issue with the triplet loss is the parameterization of $\alpha_\text{margin}$. We can overcome this problem by using the Neighbourhood Components Analysis (NCA)~\cite{roweis2004neighbourhood}. Thus, we can write the loss function using NCA as
\begin{equation}
\mathcal{L}_\text{NCA}=\frac{\exp(-\norm{f(X)-f(X^+)}^2)}
{\sum_{X^- \in C}\exp(-\norm{f(X)-f(X^-)}^2)}
\label{eq:nca}
\end{equation} 
where $C$ represents all categories except for that of the positive sample.


In the ideal scenario, when iterating over triplets of samples, we expect that the samples from the same category will be grouped in the same cluster in the embedding space. 
However, it has been shown that most of the formed triplets are not informative and visiting all possible triplet combinations is infeasible.
Therefore, the model will be trained with only a few informative triplets~\cite{rippel2016metric,schroff2015facenet,sohn2016improved}. 
An intuitive solution can be formulated by selecting those negative samples that are hard to distinguish (hard negative mining), although searching for a hard negative sample in a motion sequence dataset is computationally expensive. 
Another issue linked with the use of triplet loss is that, during a single update, the positive and negative samples are evaluated only in terms of their relative position in the embedding: thus, samples can end up close to other categories~\cite{sohn2016improved}. We address the aforementioned issue by pushing/pulling the cluster distributions instead of pushing/pulling individual samples by means of a novel loss function, dubbed MMD-NCA and described next, that is based on the distribution differences of the categories. 

\subsection{MMD-NCA} 
\label{sec:mmd-nca}


Assuming that given two different distributions $p$ and $q$, the general formulation of MMD measures the distance between $p$ and $q$ while taking the differences of the mean embeddings in Hilbert spaces, written as
\begin{align}
\mmd[k,p,q]^2 
\!=\! \|\mu_{q}-\mu_{p}\|^2  
\!=\! E_{x,x'}[k(x,x')] \!-\! 2E_{x,y}[k(x,y)] \!+\! E_{y,y'}[k(y,y')]
\label{eq:mmd_expected}
\end{align} 
where $x$ and $x'$ are drawn IID from $p$ while $y$ and $y'$ are drawn IID from $q$, and $k$ represents the kernel function
\begin{equation} 
k(x, x^{'}) = \sum_{q=1}^{K}k_{\sigma_{q}}(x,x^{'})
\label{eq:kernel}
\end{equation} 
where $k_{\sigma_{q}}$ is a Gaussian kernel with bandwidth parameter $\sigma_{q}$, while $K$ (number of kernels) is a hyperparameter.
If we replace the expected values from the given samples, we obtain 
\begin{equation}
\mmd[k,X,Y]^2 \!= \!
\frac{1}{m^2}\sum_{i=1}^{m}\sum_{j=1}^{m}k(x_{i},x_{j}') 
- \frac{2}{mn}\sum_{i=1}^{m}\sum_{j=1}^{n}k(x_{i},y_{j}) 
+ \frac{1}{n^2}\sum_{i=1}^{n}\sum_{j=1}^{n}k(y_{i},y_{j}')
\label{eq:mmd_samples}
\end{equation} 
%
where $X := \{x_{1}, x_{2}, \dots x_{m}\}$ is the sample set from $p$ and $Y := \{y_{1}, y_{2}, \dots y_{n}\}$ is the sample set from $q$. 
Hence, \eqref{eq:mmd_samples} allows us to measure the distance between the distribution of two sets.





We formulate our loss function in order to force the network to decrease the distance between the distribution of the anchor samples and that of the positive samples, while increasing the distance to the distribution of the negative samples. 

Therefore, we can rewrite \eqref{eq:nca} for a given number $N$ of anchor-positive sample pairs as
$\{(X_{1},X_{1}^{+}), (X_{2},X_{2}^{+}), \dots, (X_{N},X_{N}^{+})\}$ and $N \times M$ negative samples from the $M$ different categories $C = \{c_{1}, c_{2}, \dots, c_{M}\}$ as $\{X_{c_{1},1}^{-}, X_{c_{1},2}^{-}, \dots, X_{c_{1},N}^{-}, \dots, X_{c_{M},N}^{-} \}$; then,
\begin{equation}
\mathcal{L}_\text{MMD-NCA}
=\frac{\exp(-\mmd[k,f(X),f(X^{+})])}
{\sum_{j=1}^{M}\exp(-\mmd[k,f(X),f(X_{c_{j}}^{-})])}
\label{eq:mmdt}
\end{equation} 
%
%
%
where $X$ and $X^{+}$ represent motion samples from the same category, while $X_{c_{j}}$ represents samples from category $c_{j} \in C$.
Our single update contains $M$ different negative classes randomly sampled from the training data. 

Since the proposed MMD-NCA loss minimizes the overlap between different category distributions in the embedding while keeping the samples from the same distribution as close as possible, 
we believe it is more effective for our task than the triplet loss. We demonstrate this quantitatively and qualitatively in \secref{sec:experiments}.

\section{Network architecture} 
\label{sec:method-network}

\begin{figure}[t]
	\centering
	\includegraphics[width=1.0\columnwidth]{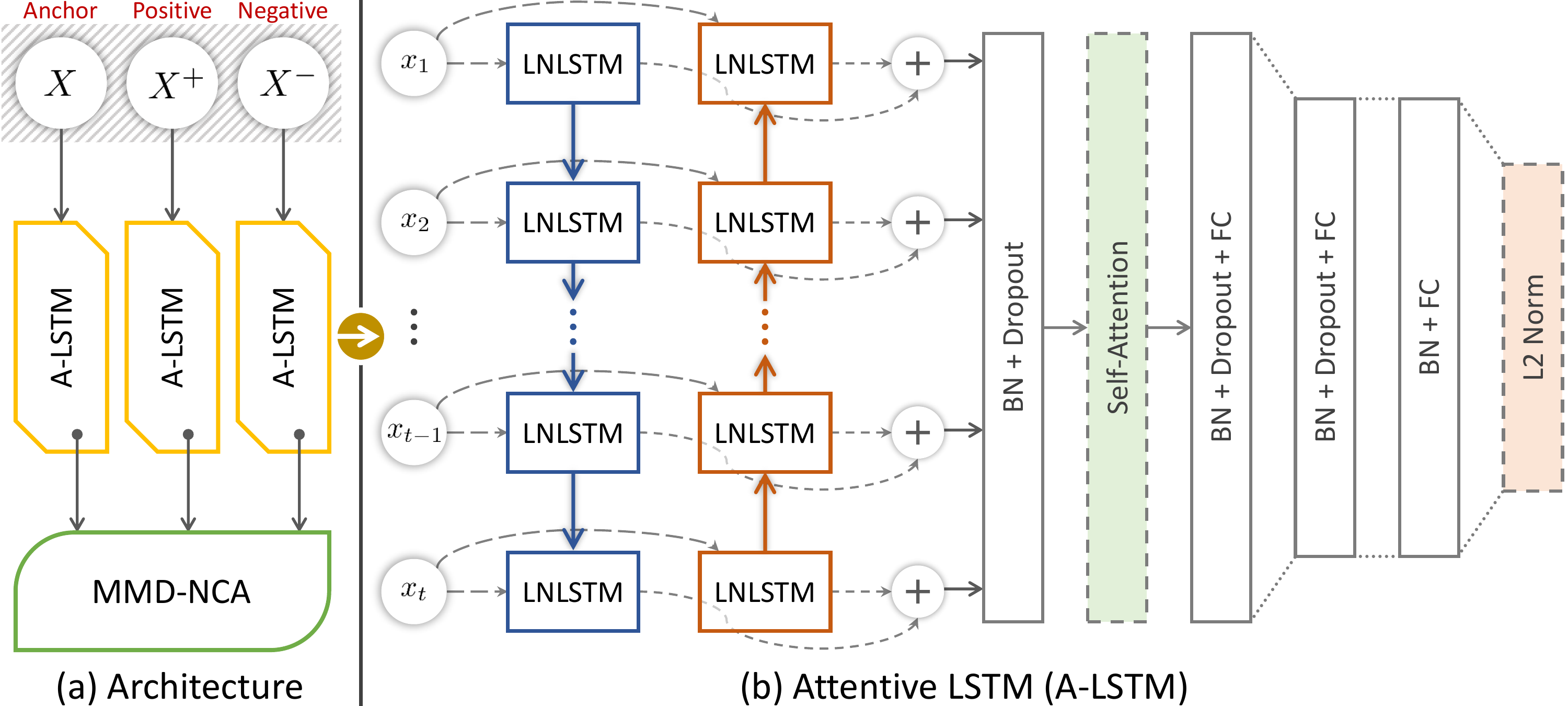}
	\setlength{\abovecaptionskip}{\RemoveAboveCaption}
	\setlength{\belowcaptionskip}{\RemoveBelowCaption}
	\caption{(a)~The proposed architecture for sequence distance learning. (b)~The proposed attention-based model that uses layer normalization.}	
	\label{fig:seqatt}
\end{figure}

Our architecture is illustrated in \figref{fig:seqatt}.
This model has two main parts: the bidirectional long short-term memory (BiLSTM)~\cite{hochreiter1997long} and the self-attention mechanism. The reason for using the long short-term memory (LSTM)~\cite{hochreiter1997long} is to overcome the vanishing gradient problem of the recurrent neural networks. In \cite{graves2013speech,greff2017lstm}, they show that LSTM can capture long term dependencies.
In the next sections, we briefly describe the layer normalization mechanism and attention mechanism that used in our architecture.

\subsection{Layer normalization} 
\label{sec:lnLSTM}

In \cite{chopra2005learning,HardNet2017,movshovitz2017no,song2016deep}, they have shown that batch normalization plays a fundamental role on the triplet model's accuracy. However, its straightforward application to LSTM architectures can decrease the accuracy of model \cite{laurent2016batch}. Due to this, we used the layer normalized LSTM~\cite{ba2016layer}. 

Suppose that $n$ time steps of motion $X=(x_{1}, x_{2}, \dots, x_{n})$ are given, then the layer normalized LSTM is described by

\begin{align}
\fb_t &= \sigma(\Wb_{fh} \hb_{t-1} + \Wb_{fx} x_t + \bb_f) \label{eq:lstmbegin}\\
\ib_t &= \sigma(\Wb_{ih} \hb_{t-1} + \Wb_{ix} x_t + \bb_i) \\
\ob_t &= \sigma(\Wb_{oh} \hb_{t-1} + \Wb_{ox} x_t + \bb_o) \\
\tilde{\cb}_t &= \tanh(\Wb_{ch} \hb_{t-1} + \Wb_{cx} x_t + \bb_c) \label{eq:lstmsimple} \\
\cb_t &= \fb_t \odot \cb_{t-1} + \ib_t \odot \tilde{\cb}_t \label{eq:lstmupdate} \\
\mb_t &=\frac{1}{H}\sum_{j}^{H} \cb^{j}_t  \label{eq:lstmmean}, \vb_t =\sqrt{\frac{1}{H}\sum_{j}^{H} (\cb^{j}_t-m_{t})^{2}} \\
\hb_t &= \ob_t \odot \tanh(\frac{\gamma_{t}}{v_{t}} \odot (\cb_t-m_{t})+\beta) \label{eq:lstmoutput}
\end{align} 
where $c_{t-1}$ and $h_{t-1}$ denotes the cell memory and cell state which comes from the previous time steps, $x_{t}$ denotes the input human pose at time $t$. $ \sigma(\cdot)$ and $\odot$ represent the element-wise sigmoid function and multiplication respectively, and $H$ denotes the number of hidden units in LSTM. 
The parameters $W_{\cdot,\cdot}$, $\gamma$ and $\beta$ are learned while $\gamma$ and $\beta$ has the same dimension of $\hb_t$. Contrary to the standard LSTM, the hidden state $\hb_t$ is computed by normalizing the cell-memory $\cb_t$. 

\subsection{Self-attention mechanism} 
\label{attention}

Intuitively, in a sequence of human motion, some poses are more informative than others. 
Therefore, we use the recently proposed self-attention mechanism \cite{lin2017structured} to assign a score for each pose in a motion sequence. 
Specifically, assuming that the sequence of states $S =\{h_{1}, h_{2}, \dots, h_{n}\}$ computed from a motion sequence $X$ that consists of $n$ time steps with 
\eqref{eq:lstmbegin} to \eqref{eq:lstmoutput}, we can effectively compute the scores for each of them by 
%
\begin{align}
r =W_{s2}\tanh(W_{s1}S^\top) 
~~~~\text{and}~~~~
a_{i} = -\log\left(\frac{\exp(r_{i})}{ \sum_{j} \exp(r_{j}) }\right) \label{eq:attn}
\end{align} 
%
%
where $r_{i}$ is $i$-th element of the $r$ while $W_{s1}$ and $W_{s2}$ are the weight matrices in $R^{k{\times}l}$ and $R^{l{\times}1}$, respectively. $a_{i}$ is the assigned score $i$-th pose in the sequence of motion. 
Thus, the final embedding $E$ can be computed by multiplying the scores $A=[a_{1}, a_{2}, \dots, a_{n}]$ and $S$, written as $E = AS$.
%
%
Note that the final embedding size only depends on the number of hidden states in the LSTM and $W_{s2}$. 
This allows us to encode the varying size LSTM outputs to a fixed sized output. 
More information about the self-attention mechanism can be found in \cite{lin2017structured}.

\section{Implementation details}
\label{sec:implement}

We use the TensorFlow framework~\cite{tensorflow2015-whitepaper} for all deep metric models that are described in this paper. 
Our model has three branches as shown in \figref{fig:seqatt}. 
Each branch consists of an attention based bidirectional layer normalized LSTM (LNLSTM) (see \secref{sec:lnLSTM}). Bidirectional LNLSTM follows a forward and backward passing of the given sequence of motion. We then denote $s_{t} =\left[s_{t,f},s_{t,b}\right]$ such that 
$s_{t,f} =\overrightarrow\lnlstm(w_{t},x_{t})$ for $t \in \left[0,N \right]$
and
$s_{t,b} =\overleftarrow\lnlstm(w_{t},x_{t})$ for $t \in \left[N,0 \right]$.
%

Given $n$ time steps of a motion sequence $X$, we compute $S=(s_{1}, s_{2}, \dots, s_{n})$ 
where $s_{t}$ is the concatenated output of the backward and forward pass of the LNLSTM which has 128 hidden units. 
The bidirectional LSTM is followed by the dropout and the standard batch normalization. The output of the batch normalization layer is forwarded to the attention layer (see \secref{attention}), which produces the fixed size of the output. 
The attention layer is followed by the structure: \{FC(320,), dropout, BN, FC(320), BN, FC(128), BN, $l_{2}$ Norm\}, where FC($m$) means fully connected layer with $m$ as the hidden units and BN means batch normalization. 
All the FC layers are followed by the rectified linear units except for the last FC layer. The self-attention mechanism is derived from the implementation of \cite{lin2017structured}. 
Here, the $W_{s1}$ and $W_{s2}$ parameters from \eqref{eq:attn} have the dimensionality of $R^{200{\times}10}$ and $R^{10{\times}1}$, respectively. 
We use the dropout rate of 0.5. The same dropout mask is used in all branches of the network in \figref{fig:seqatt}. 
In our model, all squared weight matrices are initialized with random orthogonal matrices while the others are initialized with uniform distribution with zero mean and 0.001 standard deviation. 
The parameters $\gamma$ and $\beta$ in \eqref{eq:lstmoutput} are initialized with zeros and ones, respectively. 

\paragraph{Kernel designs.} 

The MMD-NCA loss function is implicitly associated with a family of characteristic kernels. 
Similar to the prior MMD papers \cite{li2015generative,sutherland2016generative}, we consider a mixture of $K$ radial basis functions in \eqref{eq:kernel}. We fixed $K = 5$ and $\sigma_{q}$ to be ${1, 2, 4, 8, 16}$.

\paragraph{Training.}

Our single batch consists of randomly selected categories where each category has 25 samples. We selected 5 category as negative.
Although the MMD \cite{gretton2012kernel} metric requires a high number of samples to understand the distribution moments, we found that 25 is sufficient for our tasks. 
Training each batch takes about 10 seconds on a Titan X GPU. 
All the networks are trained with 5000 updates and they all converged before the end of training. 
During training, analogous to the curriculum learning, we start training on the samples without noise and then added Gaussian noise with zero mean and increasing standard deviation. 
We use stochastic gradient descent with the moment as an optimizer for all models. 
The momentum value is set to 0.9, and the learning rate started from 0.0001 with an exponential decay of 0.96 every 50 updates. 
We clip the whole gradients by their global norm to the range of $-$25 and 25.

\section{Experimental results} \label{sec:experiments}

We compare our MMD-NCA loss against the methods from DTW~\cite{vintsyuk1968speech}, MDDTW~\cite{mei2016learning}, CTW~\cite{zhou2009canonical} and  GDTW~\cite{zhou2016generalized},  as well as four state-of-the-art deep metric learning approaches: DCTW~\cite{trigeorgis2018deep}, triplet~\cite{schroff2015facenet}, triplet+GOR~\cite{zhang2017learning}, and the $N$-Pairs deep metric loss~\cite{gretton2012kernel}. 
Primarily, these methods are evaluated through action recognition task in \secref{sec:action_rec}.
 In order to look closely into the performance of this evaluation, we analyze the actions retrieved by the proposed method in the same section and the contribution of the self-attention mechanism from \secref{attention} into the algorithm in \secref{sec:attention_vis}. 
Since one of the datasets~\cite{carnegie55} labeled the actions with their corresponding subjects, we also investigate the possibility of performing a person identification task wherein, instead of measuring the similarity of the pose, we intend to measure the similarity the actors themselves based on their movement. 
To have a fair comparison, we only used our attention based LSTM architecture for all methods and only changed the loss function except the DCTW~\cite{trigeorgis2018deep}. Prosed loss function in DCTW~\cite{trigeorgis2018deep} requires the two sequences, therefore we remove the attention layer and use only our LSTM model.
Notably, all deep metric learning methods are evaluated and trained with the same data splits.

\paragraph{Performance Evaluation.}

We follow the same evaluation protocol as defined in \cite{song2016deep,zhang2017learning}.
All models are evaluated for the clustering quality and false positive rate (FPR) on the same test set which consists of unseen motion categories. 
We compute the FPR for 90\%, 80\% and 70\% true positive rates.
In addition, we also use the Normalized Mutual Information measure (NMI) and \fone score to measure the cluster quality where the NMI is the ratio between mutual information and sum of class and cluster labels entropies while the \fone score is the harmonic mean of precision and recall.

\paragraph{Datasets and Pre-processing.}

We tested the models on two different datasets: (1)~the CMU Graphics Lab motion capture database (CMU mocap)~\cite{carnegie55}; and, (2)~the Human3.6M dataset~\cite{Ionescu2014}. 
The former~\cite{carnegie55} contains 144 different subjects where each subject performs natural motions such as \emph{walking}, \emph{dancing} and \emph{jumping}. Their data is recorded with the mocap system and the poses are represented with 38 joints in 3D space. 
Six joints are excluded because they have no movement. 
We align the poses with respect to the torso and, to avoid the gimbal lock effect, the poses are expressed in the exponential map~\cite{taylor2007modeling}. 
Although the original data runs at 120Hz with different lengths of motion sequences, we down-sampled the data to 30Hz during training and testing.

Furthermore, the Human3.6M dataset~\cite{Ionescu2014} consists of 15 different actions and each action was performed by seven different professional actors. 
The actions are mostly selected from daily activities such as \emph{walking}, \emph{smoking}, \emph{engaging in a discussion}, \emph{taking pictures} and \emph{talking on the phone}. 
We process the dataset in the same way as the same as CMU mocap. 

\definecolor{Res6}{rgb}{0.93,0.93,0.93}
\definecolor{Res5}{rgb}{0.99,0.94,0.91}
\definecolor{Res3}{rgb}{0.93,0.96,0.91}
\definecolor{Res4}{rgb}{0.93,0.95,0.98}
\definecolor{Res2}{rgb}{0.95,0.95,0.95}
\definecolor{Res1}{rgb}{1.0,0.97,0.90}

\begin{table*}[t]
	\centering
	\begin{tabular}{
	>{\raggedright\arraybackslash}p{2.75cm}|
	>{\columncolor{Res1}}>{\centering\arraybackslash}p{1.4cm}|
	>{\columncolor{Res1}}>{\centering\arraybackslash}p{1.4cm}|
	>{\columncolor{Res1}}>{\centering\arraybackslash}p{1.4cm}|
	>{\columncolor{Res2}}>{\centering\arraybackslash}p{1.4cm}|
	>{\columncolor{Res2}}>{\centering\arraybackslash}p{1.4cm}|
	>{\columncolor{Res2}}>{\centering\arraybackslash}p{1.4cm}} 
			\multicolumn{1}{c}{} & 
			\multicolumn{3}{c}{\small {CMU}} &   
			\multicolumn{3}{c}{\small {Human3.6M}} \\
		\cmidrule{2-7}
		\multicolumn{1}{c}{}
		& \multicolumn{1}{c}{FPR-90}  &  \multicolumn{1}{c}{FPR-80} & \multicolumn{1}{c|}{FPR-70}& \multicolumn{1}{c}{FPR-90}& 
		\multicolumn{1}{c}{FPR-80}& \multicolumn{1}{c}{FPR-70}\\
		\midrule		
DTW \cite{vintsyuk1968speech}	&	 47.98 	&	 42.92 	&	 37.62	&	 49.64 	&	 47.96 	&	 44.38	\\
MDDTW \cite{mei2016learning} 	&	 44.60 	&	 39.07 	&	 34.04	&	 49.72 	&	 45.87 	&	 44.51	\\
CTW \cite{zhou2009canonical} 	&	 46.02 	&	 40.96 	&	 39.11	&	 47.63 	&	 43.10 	&	 42.18\label{key}	\\
GDTW \cite{zhou2016generalized} 	&	 45.61 	&	 39.95 	&	 35.24	&	 46.06 	&	 42.72 	&	 40.04	\\
DCTW \cite{trigeorgis2018deep} 	&	 40.56 	&	 38.83 	&	 26.95	&	 41.39 	&	 39.18 	&	 36.71	\\
Triplet \cite{schroff2015facenet} 	&	 39.72 	&	 33.82 	&	 28.77	&	 42.78 	&	 40.15 	&	 36.01	\\
Triplet + GOR \cite{zhang2017learning} 	&	 40.32 	&	 33.97 	&	 27.78	&	 42.03 	&	 37.61 	&	 33.95	\\
N-Pair \cite{sohn2016improved} 	&	 40.11 	&	 32.35 	&	 26.16	&	 40.46 	&	 39.56 	&	 36.52	\\
MMD-NCA (\emph{Ours}) 	&	 \textbf{32.66} 	&	 \textbf{25.66} 	&	 \textbf{20.29}	&	 \textbf{38.42} 	&	 \textbf{36.54} 	&	 \textbf{33.13}	\\
		\midrule															
-- without Attention  	&	 41.22 	&	 35.36 	&	 30.04	&	 45.03 	&	 42.07 	&	 41.01	\\
-- without LN  	&	 37.27 	&	 30.21 	&	 27.95	&	 44.25 	&	 41.69 	&	 38.09	\\
-- Linear Kernel  	&	 39.80 	&	 33.92 	&	 29.00	&	 46.35 	&	 41.68 	&	 37.69	\\
-- Polynomial Kernel 	&	 36.80 	&	 30.35 	&	 24.98	&	 43.60 	&	 40.03 	&	 35.62	\\

		\bottomrule
	\end{tabular}
	\setlength{\belowcaptionskip}{\RemoveBelowCaption}
	\caption{
		False positive rate of action recognition for CMU mocap and Human3.6M datasets. \label{tbl:cmu-action}} 
\end{table*}

\begin{figure} [!ht]
	\centering
	\includegraphics[width=1.00\textwidth]{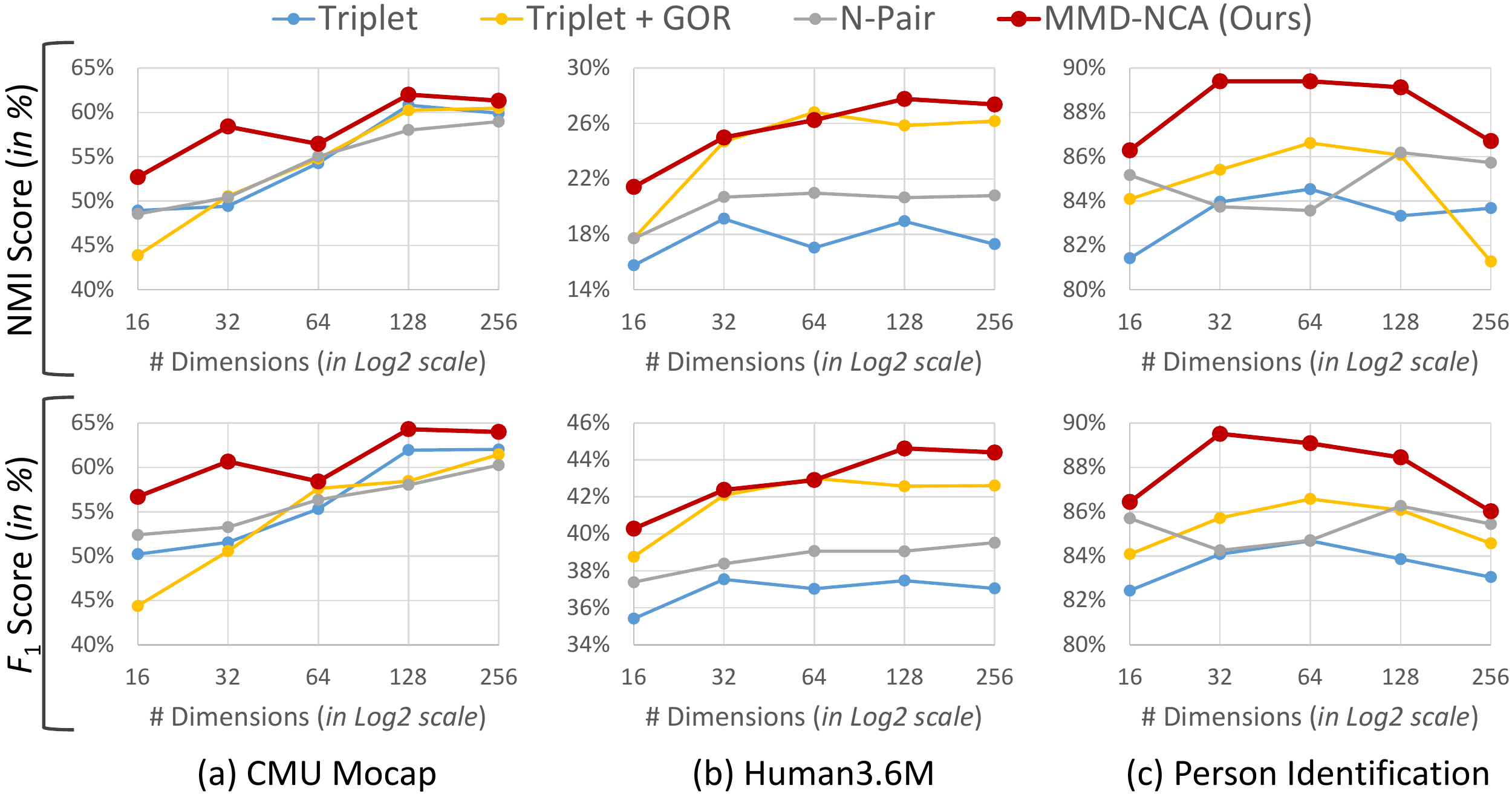}
	\setlength{\abovecaptionskip}{\RemoveAboveCaption}
	\setlength{\belowcaptionskip}{\RemoveBelowCaption}
	\caption{NMI and \fone score for the action recognition task using the (a)~CMU Mocap and (b)~Human3.6M datasets; and, (b) for person identification task. }
	\label{fig:nmi_f1_score}
\end{figure}

\subsection{Action recognition} 
\label{sec:action_rec}

In this experiment, we tested our model on both the CMU mocap~\cite{carnegie55} and the Human3.6M~\cite{Ionescu2014} datasets for unseen motion categories. 
We categorize the CMU mocap dataset into 38 different motion categories where the motion sequences which contain more than one category are excluded. 
Among them, we selected 19 categories for training and 19 categories for testing. For the Human3.6M~\cite{Ionescu2014}, we used all the given categories, and selected 8 categories for training and 7 categories for testing. 

Although our model allows us to train with varying sizes of motion sequence, we train with a fixed size, since varying sizes slow down the training process. 
We divided the motion sequences into 90 consecutive frames (\ie approximately 3 seconds) and leave a gap of 30 frames. 
However, at test time, we divided the motion sequences only if it is longer than 5 seconds by leaving a 1-second gap; otherwise, we keep the original motion sequence. 
We found this processing effective since we observe that, in sequence of motions longer than 5 seconds, the subjects usually repeat their action. 
We also consider training without clipping but it was not possible with available the GPU resources.

\begin{figure} [t]
	\centering
	\includegraphics[width=1.00\textwidth]{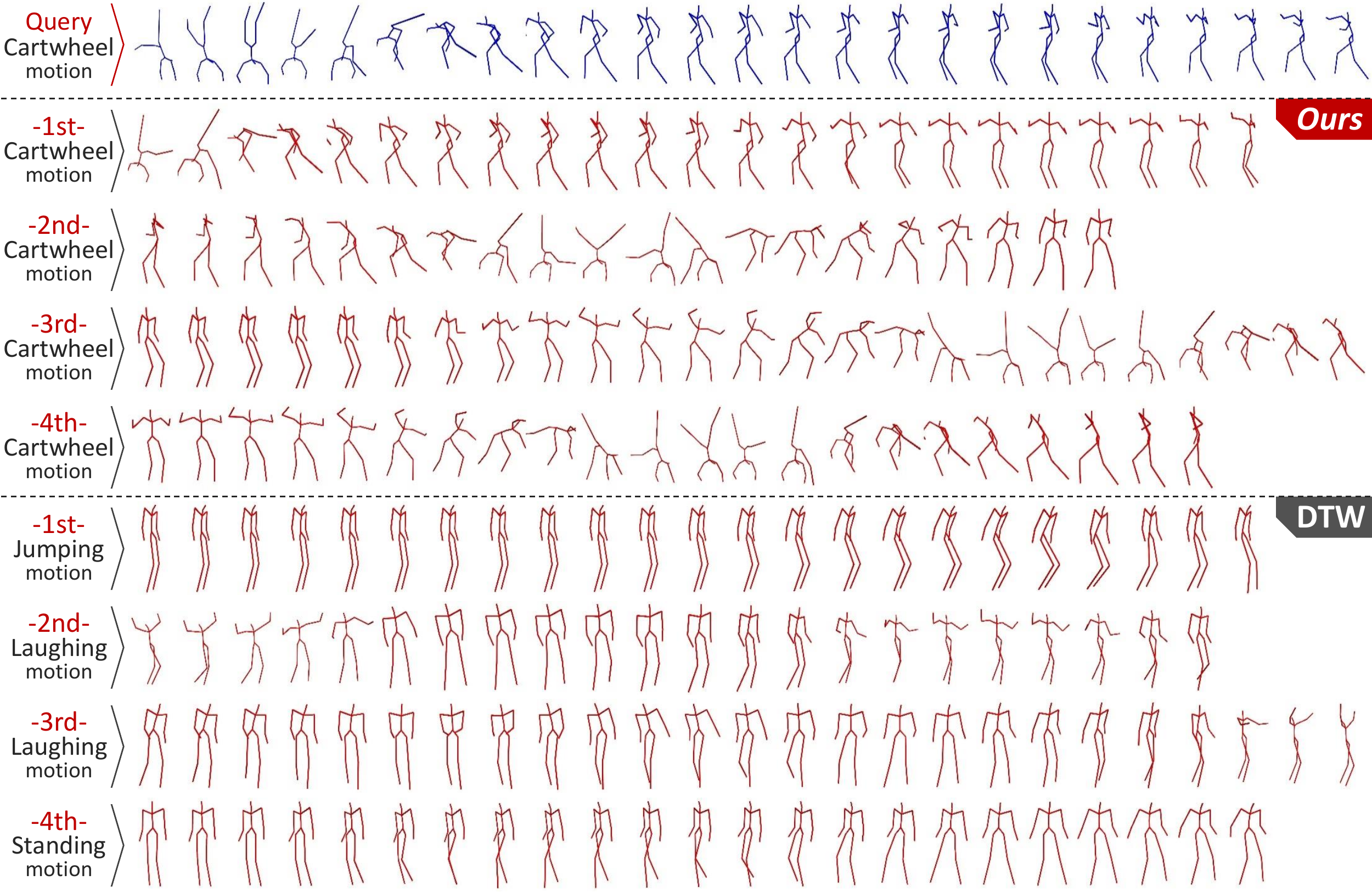}
	\setlength{\abovecaptionskip}{\RemoveAboveCaption}
	\setlength{\belowcaptionskip}{\RemoveBelowCaption}
	\caption{Comparison of cartwheel motion query on the CMU mocap dataset between our approach and DTW~\cite{vintsyuk1968speech}. The motion in the first row is query and the rest are four nearest neighbors for each method, which are sorted by the distance.}
	\label{fig:query_ours}
\end{figure}

\paragraph{False Positive Rate.}
The FPR at different percentages on CMU mocap and Human3.6M are reported in \tabref{tbl:cmu-action}. 
With a true positive rate of 70\%, the  learning approaches~\cite{schroff2015facenet,trigeorgis2018deep,zhang2017learning,sohn2016improved} including our approach achieve up to 17\% improvement in FPR relative to DTW~\cite{vintsyuk1968speech}, MDDTW~\cite{mei2016learning}, CTW~\cite{zhou2009canonical} and  GDTW~\cite{zhou2016generalized}.
%
Moreover, our approach further improves the results up to 6\% and 0.8\% for CMU mocap and Human3.6m datasets, respectively, against the state-of-the-art deep learning approaches~\cite{schroff2015facenet,trigeorgis2018deep,zhang2017learning,sohn2016improved}.

\paragraph{NMI and \fone score.}
\figref{fig:nmi_f1_score}(a) plots the NMI and \fone score with varying size of embedding for the CMU mocap dataset. 
In both the NMI and \fone metrics, our approach produces the best clusters at all the embedding sizes. 
Compared to other methods, the proposed approach is less sensitive to the changes of the embedding size. 
Moreover, \figref{fig:nmi_f1_score}(b) illustrates the NMI and \fone score on Human3.6M dataset where we observe similar performance as the CMU mocap dataset and acquire the best results.

\paragraph{Action retrieval.}
In order to investigate further, we query a specific motion from the CMU mocap test set, and compare the closest action sequences that our approach and DTW~\cite{vintsyuk1968speech} retrieve based on their respective similarity measure. 
In \figref{fig:query_ours}, we demonstrate this task as we query the challenging cartwheel motion (see first row).
Our approach successfully retrieves the semantically similar motions sequences, despite the high variation on the length of sequences. On the other hand, DTW~\cite{vintsyuk1968speech} fails to match the query to the dataset because the distinctive pose appears on a small portion of the sequence. This implies that the large portion, where the actor stands, dominates the similarity measure. Note that we do not have the same problem due to the self-attention mechanism from \secref{attention} (see \secref{sec:attention_vis} for the evaluation).

%

\subsection{Person identification}
\label{sec:person_id_eval}

Since the CMU mocap dataset also includes the specific subject associated to each motion, we explore the potential application of person identification. 
In contrast to the action recognition and action retrieval from \secref{sec:action_rec} where the similarity measure is calculated based on the motion category, this task tries to measure the similarity with respect the actor. 
In this experiment, we construct the training and test set in the same way as \secref{sec:action_rec}.
We included the subjects which have more than three motion sequences, which resulted in 68 subjects.
Among them, we selected 39 subjects for training and the rest of the 29 subjects for testing. 

\tabref{tbl:cmu_gait} shows the FPR for the person identification task for varying percentages of true positive rate with embedding size of 64. 
Here, all deep metric learning approaches including our work significantly improve the accuracy against the DTW, MDDTW, CTW and  GDTW. 
Overall, our method outperforms all the approaches for all FPR with a 20\% improvement against DTW~\cite{vintsyuk1968speech}, MDDTW~\cite{mei2016learning}, CTW~\cite{zhou2009canonical} and  GDTW~\cite{zhou2016generalized}, and a 2\% improvement compared to the state-of-the-art deep learning approaches~\cite{schroff2015facenet,trigeorgis2018deep,zhang2017learning,sohn2016improved}. 
Moreover, when we evaluate the NMI and the \fone score for the clustering quality in different embedding sizes, \figref{fig:nmi_f1_score}(c) demonstrates that our approach obtains the state-of-the-art results with a significant margin.



	
	

\begin{table*}[t]
	\centering
	\begin{tabular}{
	>{\raggedright\arraybackslash}p{2.75cm}|
	>{\centering\arraybackslash}p{1.4cm}|
	>{\centering\arraybackslash}p{1.4cm}|
	>{\centering\arraybackslash}p{1.4cm}|
	>{\centering\arraybackslash}p{1.4cm}|
	>{\centering\arraybackslash}p{1.4cm}|
	>{\centering\arraybackslash}p{1.4cm}} 
		\toprule 	
		\multicolumn{1}{c}{}
		& \multicolumn{1}{c}{FPR-95}  &  \multicolumn{1}{c}{FPR-90} & \multicolumn{1}{c}{FPR-85}& \multicolumn{1}{c}{FPR-80}& 
		\multicolumn{1}{c}{FPR-75}& \multicolumn{1}{c}{FPR-70}\\
		\midrule		
		DTW \cite{vintsyuk1968speech}& 46.22 & 43.19 & 38.70 & 32.36 & 27.61  & 22.85\\
		MDDTW \cite{mei2016learning} & 49.67 & 45.89 & 40.36 & 35.46 & 31.69  & 28.44\\
		CTW \cite{zhou2009canonical} & 45.23 & 40.14 & 35.69 & 29.50 & 25.91  & 20.35\\
		GDTW \cite{zhou2016generalized} & 44.65 & 40.54 & 35.03 & 28.07 & 24.31  & 19.32\\
		DCTW \cite{trigeorgis2018deep} & 32.45 & 20.24 & 18.15 & 15.91 & 13.78  & 10.31\\
		Triplet \cite{schroff2015facenet} & 22.58 & 18.13 & 11.30 & 9.63 & 8.36  & 6.51\\
		Triplet + GOR \cite{zhang2017learning} & 28.37 & 16.69 & 10.27 & 8.64 & 7.28 & 4.38\\
		N-Pair \cite{sohn2016improved} & 22.84 & 15.31 & 8.94 & 5.69 & 4.82 & 4.56\\
		MMD-NCA (\emph{Ours}) & \textbf{19.31} & \textbf{10.42} & \textbf{8.26} & \textbf{5.62} & \textbf{3.91} & \textbf{2.55}\\
\midrule
		-- without Attention  & 36.10 & 26.15 & 22.48 & 20.94 & 19.21 & 16.78\\
		-- without LN  & 26.63 & 18.43 & 12.81 & 10.27 & 8.58 & 7.36\\
		-- Linear Kernel  & 35.75 & 30.97 & 25.93 & 15.13 & 11.93 & 10.42\\
		-- Polynomial Kernel & 27.25 & 21.18 & 17.91 & 10.93 & 8.97 & 5.93\\
		\bottomrule
	\end{tabular}
	\setlength{\belowcaptionskip}{\RemoveBelowCaption}
	\caption{False positive rate of person identification for CMU mocap dataset. \label{tbl:cmu_gait}} 
\end{table*}

\subsection{Attention visualization}
\label{sec:attention_vis}

\begin{figure} [t]
	\centering
	\includegraphics[width=1.00\textwidth]{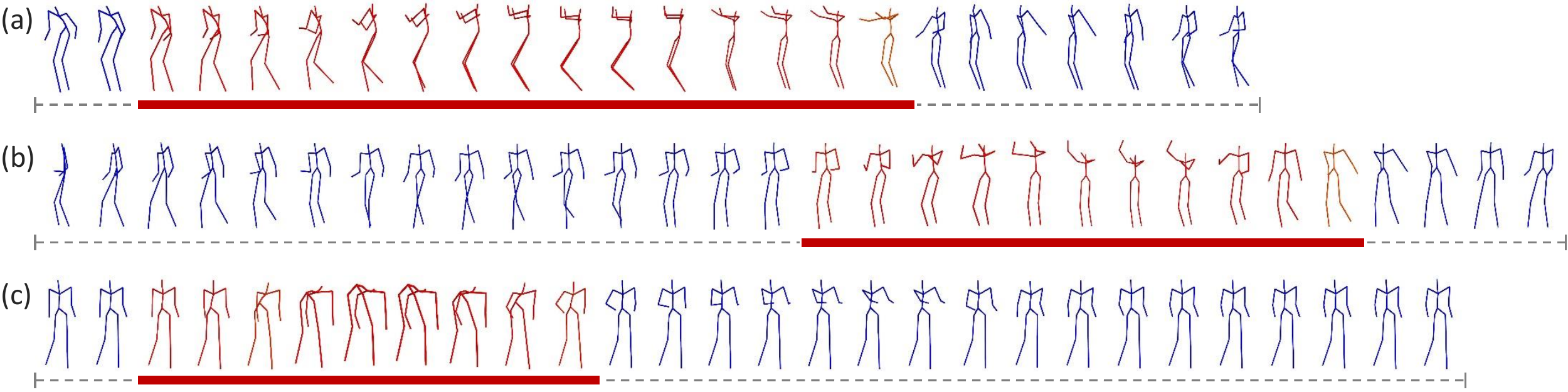}
	\setlength{\abovecaptionskip}{\RemoveAboveCaption}
	\setlength{\belowcaptionskip}{\RemoveBelowCaption}
	\caption{Attention visualization: the poses in red show where the model mostly focused its attention. Specifically, we mark as red those frames associated with each column-wise global maximum in $A$, together with the previous and next 2 frames. For visualization purposes, the sequences are subsampled by a factor of 4. }
	\label{fig:visatt}
\end{figure}

The objective of the self-attention mechanism from \secref{attention} is to focus on the poses which are the most informative about the semantics of the motion sequence. 
Thus, we expect our attention mechanism to focus on the descriptive poses in the motion, which allows the model to learn more expressive embeddings. 
Based on the peaks of $A$ which is composed of $a_i$ from \eqref{eq:attn}, we illustrate this behavior in \figref{fig:visatt} where the first two rows belong to the basketball sequence while the third belong to the bending sequence. Notably, all the sequences have different lengths.

Despite the variations in the length of the motion, the model focuses when the actor throws the ball which is the most informative part of the motion for \figref{fig:visatt}(a-b); while, for the bending motion in \figref{fig:visatt}(c), it also focuses on the distinctive regions of the motion sequence.
Therefore, this figure illustrate that the self-attention mechanism
successfully focuses on the most informative part of the sequence. 
This implies that 
the model discards the non-informative parts in order to embed long motion sequences to a low dimensional space without losing the semantic information.

\section{Ablation study} \label{sec:compare}
We evaluate our architecture with different configurations to better appreciate each of our contributions separately. All models are trained with MMD-NCA loss and with an embedding of size 128. 
Tables \ref{tbl:cmu-action} and \ref{tbl:cmu_gait} show the effect of the layer normalization~\cite{ba2016layer}, the self-attention mechanism~\cite{lin2017structured} and the kernel selection in terms of FPR. We use the same architecture for linear, polynomial, and MMD-NCA and only change the kernel function in \eqref{eq:kernel}.
Notably, the removal of the self-attention mechanism yields the biggest drop in NMI and \fone on all the datasets. In addition, Both the layer normalization and the self-attention improve the resulting FPR by 7\% and 10\%, respectively. In terms of kernel selection, the results shows that selecting the kernel which takes into account higher moments yields better results. Comparing the two tasks, the person identification is the one that benefits from our architecture the most.  

\section{Conclusion} 
\label{sec:conclusion}

In this paper, we propose a novel loss function and network architecture to measure the similarity of two motion sequences. Experimental results on the CMU mocap~\cite{carnegie55} and Human3.6M~\cite{Ionescu2014} datasets show that our approach obtain state-of-the-art results. We also have shown that metric learning approaches based on deep learning can improve the results up to 20\% against metrics commonly used for similarity among human motion sequences. 
As future work, we plan to generalize the proposed MMD-NCA framework to time-series, as well as investigate different types of kernels.

\bibliographystyle{splncs04}
\bibliography{cvbib}
%
%
%
%
%
%
%
%
\end{document}